\documentclass{vgtc}                          

\ifpdf
  \pdfoutput=1\relax                   
  \usepackage{graphicx}                
  \DeclareGraphicsExtensions{.pdf,.png,.jpg,.jpeg} 
\else
  \usepackage{graphicx}                
  \DeclareGraphicsExtensions{.eps}     
\fi%

\graphicspath{{figures/}{pictures/}{images/}{./}} 

\usepackage{microtype}                 
\PassOptionsToPackage{warn}{textcomp}  
\usepackage{textcomp}                  
\usepackage{mathptmx}                  
\usepackage{times}                     
\usepackage{cite}                      
\usepackage{tabu}                      
\usepackage{booktabs}                  
\usepackage{amsmath}
\usepackage{amssymb}
\usepackage{colortbl}
\usepackage{xspace}
\usepackage{multirow}
\usepackage{makecell}
\usepackage{hyperref}
\hypersetup{colorlinks=true}

\onlineid{7725}

\vgtccategory{Research, Algorithm,  System}

\vgtcinsertpkg

\title{SimpleMapping: Real-Time Visual-Inertial Dense Mapping \\ with Deep Multi-View Stereo}

\author{Yingye Xin${}^{1 \, *}$
\and Xingxing Zuo${}^{1\, 2\, }$\thanks{indicates equal contribution.} \,\thanks{indicates corresponding author.}
\and Dongyue Lu${}^{1}$
\and Stefan Leutenegger${}^{1\, 2}$}
\affiliation{\scriptsize ${}^1$Smart Robotics Lab, Technical University of Munich, Germany \\
${}^2$Munich Center for Machine Learning (MCML), Germany\\
\{yingye.xin, \,xingxing.zuo, \,dongyue.lu, \,stefan.leutenegger\}@tum.de
}

\teaser{
  \centering

  \includegraphics[width=0.8\linewidth]{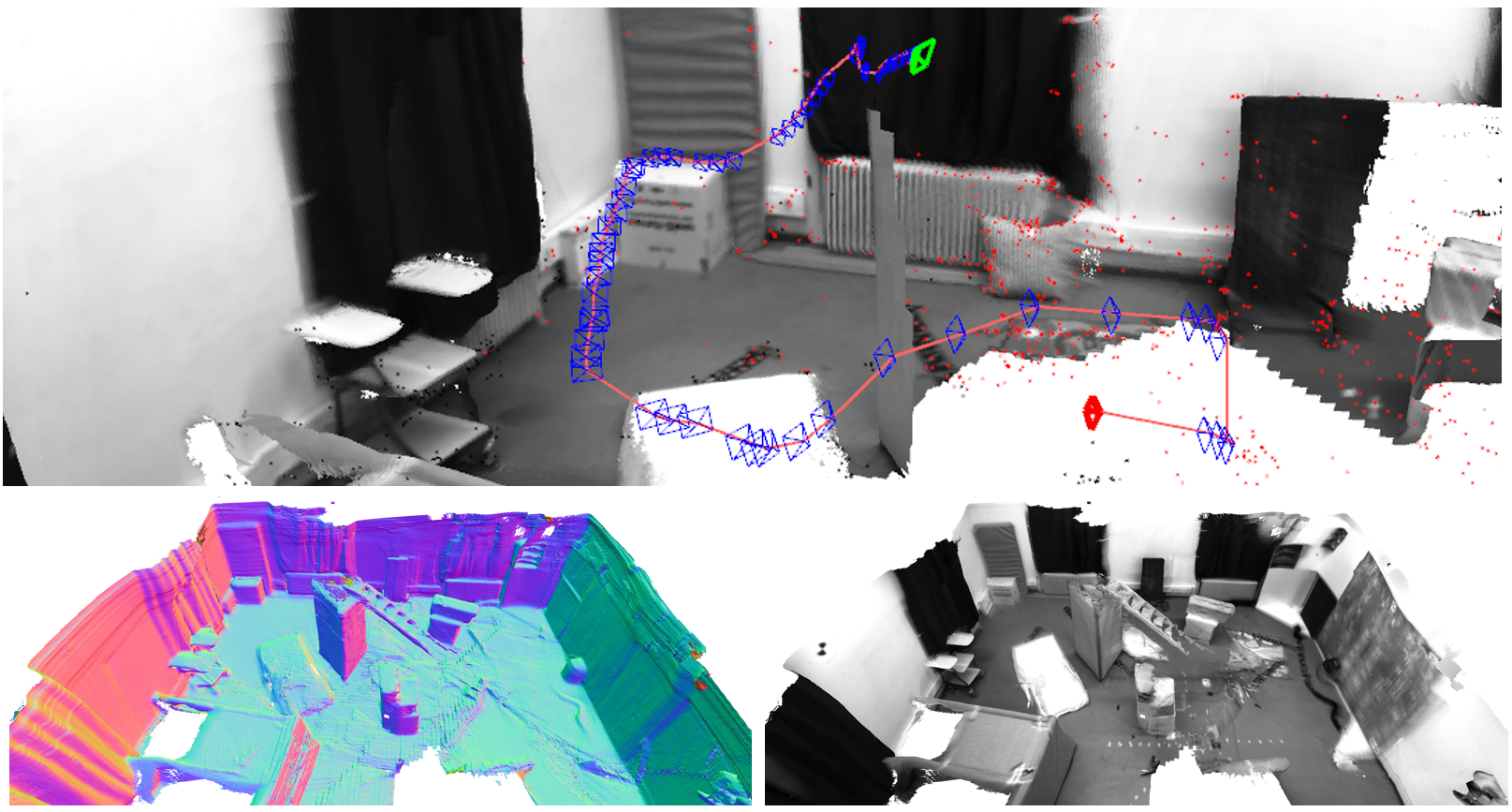}
  \vspace{-1em}
  \caption{Showcase of EuRoC/V201\cite{Burri25012016} reconstructed in real-time with SimpleMapping. As shown in the top image, the current camera frame in green is navigating across the room while the surface mesh with texture is incrementally reconstructed online. The resulting global 3D dense map is presented on the bottom right, accompanied by the corresponding normal map on the left.}
  \label{fig:cover}
}

\abstract{We present a real-time visual-inertial dense mapping method capable of performing incremental 3D mesh reconstruction with high quality using only sequential monocular images and inertial measurement unit (IMU) readings. 6-DoF camera poses are estimated by a robust feature-based visual-inertial odometry (VIO), which also generates noisy sparse 3D map points as a by-product. We propose a sparse point aided multi-view stereo neural network (SPA-MVSNet) that can effectively leverage the informative but noisy sparse points from the VIO system. The sparse depth from VIO is firstly completed by a single-view depth completion network. This dense depth map, although naturally limited in accuracy, is then used as a prior to guide our MVS network in the cost volume generation and regularization for accurate dense depth prediction. Predicted depth maps of keyframe images by the MVS network are incrementally fused into a global map using TSDF-Fusion. We extensively evaluate both the proposed SPA-MVSNet and the entire dense mapping system on several public datasets as well as our own dataset, demonstrating the system's impressive generalization capabilities and its ability to deliver high-quality 3D reconstruction online. Our proposed dense mapping system achieves a 39.7\% improvement in F-score over existing systems when evaluated on the challenging scenarios of the EuRoC dataset.
} 

\CCScatlist{
  \CCScatTwelve{Real-Time System}{Dense Mapping}{Depth Completion}{Multi-view Stereo} 
}

\nocopyrightspace

\makeatletter
\DeclareRobustCommand\onedot{\futurelet\@let@token\@onedot}
\def\@onedot{\ifx\@let@token.\else.\null\fi\xspace}

\def\ie{\emph{i.e}\onedot} 
\makeatother

\newcommand{\PAR}[1]{\vskip4pt \noindent{\bf #1~}}
\newcommand\scalemath[2]{\scalebox{#1}{\mbox{\ensuremath{\displaystyle #2}}}}

\begin{document}


\firstsection{Introduction}

\maketitle

Real-time dense mapping of the 3D environment is crucial for a variety of applications, including mixed reality, autonomous navigation for mobile robots, and 3D scanning. While dense mapping techniques based on range sensors such as LiDAR and RGB-D cameras have become the preferred method for creating dense maps, recovering 3D dense geometry with a monocular camera has more practical and feasible application scenarios. 
However, this task is challenging due to inherent ambiguities in structure estimation solely from color images, and high computational demands for simultaneously estimating poses and recovering 3D dense structures without sensory distance measurements.

Existing feature-based simultaneous localization and mapping (SLAM) systems have demonstrated that optimizing poses and discriminative sparse points (3D landmarks) jointly is crucial for achieving highly accurate estimation. The discriminative 3D sparse points can be continuously optimized in the state estimator if they are observed in sequential images. Thus, even if they are noisy, their depth estimation is more accurate and reliable compared to other image regions that are less distinctive in their texture.
Monocular visual SLAM systems are known to suffer from limitations in that they can only provide up-to-scale pose estimation and dense mapping.  Incorporating an IMU~-- sensing linear acceleration and rotation speed~-- can overcome this limitation. A commonly used and lightweight sensor suite for real-time localization and mapping is the combination of a monocular camera and an IMU, which constitutes a conventional visual-inertial navigation system (VINS).

Traditional real-time monocular visual(-inertial) SLAM systems~\cite{newcombe2011dtam,pizzoli2014remode,engel2014lsd,mur2015probabilistic,yang2017real,rosinol2020kimera} perform (semi-)dense mapping by matching pixel observations of 3D points and refining their depth values based on the matched pixel observations across different views with parallel computation techniques. However, the reliance on correct pixel matches often results in incomplete and noisy 3D dense reconstruction with undesired artifacts, particularly in low-textured regions.

In recent years, several learning-based SLAM methods with real-time dense mapping capability have emerged, demonstrating impressive reconstruction results by leveraging learned priors of 3D structures. CNN-SLAM~\cite{tateno2017cnn} is one pioneering work that fuses CNN-predicted single-view dense depth maps with depth measurements recovered from direct monocular SLAM, leading to more complete dense reconstruction. 
Dense depth prediction from neural networks can be very noisy, whereas optimizing the depth values of a large number of pixels can be very computationally intensive. 
CodeSLAM~\cite{bloesch2018codeslam} and its derivatives~\cite{zuo2021codevio,matsuki2021codemapping} employ conditional variational auto-encoder (CVAE) networks to predict and encode a dense depth map from a single-view RGB image. The dense depth is encoded as latent variables in the bottleneck of the CVAE, which can be continuously optimized in the SLAM state estimator alongside the navigation states.
CodeVIO~\cite{zuo2021codevio} tightly couples the VIO and CVAE-like single-view depth prediction network by fusing sparse depth from VIO and single-view RGB image into the CVAE to predict dense depth map. The navigation states in the VIO estimator are online optimized by the CVAE-predicted dense depth measurements.
However, these methods rely on single-view depth prediction networks which inherently ignore the dense geometric cues and consistency across multi-view frames during network inference. Consequently, they fail to produce high-quality dense reconstructions, despite employing exquisite depth refinement schemes in later stages.
TANDEM~\cite{koestler2021tandem} is a real-time monocular dense mapping system that utilizes estimated poses from Direct Sparse Odometry~\cite{engel2017direct} and incrementally fuses dense depth maps predicted from a multi-view stereo network (MVS). This monocular visual odometry system can generate sound up-to-scale dense reconstruction at moderate camera motion speeds. However, its performance is notably constrained in scenarios with fast motion, like the EuRoC~\cite{Burri25012016} `difficult' sequences, due to challenges such as poor tracking of poses and algorithmic latencies.


In this paper, we propose SimpleMapping, a real-time visual-inertial dense mapping system that seamlessly integrates the poses and sparse 3D landmarks obtained from a feature-based VIO with an efficient MVS network.
To the best of our knowledge, we are among the first to demonstrate that the 3D sparse points, which may be noisy but easily obtained from traditional feature-based VIO systems, can substantially enhance the accuracy and robustness of dense depth prediction in MVS networks.
The key contributions of the proposed method can be summarized as follows:
\begin{itemize}
	\item We propose an efficient MVS network, SPA-MVSNet, that exhibits state-of-the-art dense depth prediction performance by leveraging sparse 3D points from a feature-based VIO system. We first complete the sparse depth using a lightweight single-view depth completion network. The completed rough depth is then seamlessly integrated into the MVS network for guiding the cost volume generation and regularization. This prevents the MVS from wasting computation on empty 3D space that dominates the scene.
	\item We develop a complete incremental dense mapping system that combines the MVS network with the feature-based VIO for robust and accurate pose estimation and dense reconstruction with great generalization capability.
	\item The proposed SPA-MVSNet only trained on the ScanNet dataset~\cite{dai2017scannet} is examined on both the ScanNet~\cite{dai2017scannet} test split and the 7-Scenes dataset~\cite{glocker2013real}. The full visual-inertial dense mapping system is extensively evaluated on two public datasets, EuRoC~\cite{Burri25012016} and ETH3D~\cite{schops2019bad}, as well as our own collected dataset to demonstrate its scalability to completely-unseen scenarios. The proposed method achieves state-of-the-art dense reconstruction quality at remarkably fast run-time, surpassing compared methods by a large margin. 
\end{itemize}

\label{sec:related_works}
\section{Related Works}

The proposed visual-inertial dense mapping system is highly relevant to deep multi-view stereo neural networks and dense SLAM systems. There is a rich body of literature in these related fields, and in this section, we will review only the most pertinent works.

\PAR{Geometric Dense Visual SLAM.}
Dense mapping can greatly benefit from the depth measurements from RGB-D cameras (or also LiDAR). 3D dense maps can be incrementally reconstructed by efficiently fusing the sequentially captured dense depth maps, which can be represented as point clouds, surfels, occupancy maps, or Truncated Signed Distance Functions (TSDFs)~\cite{newcombe2011kinectfusion, endres2013a, whelan2015elasticfusion, dai2017bundlefusion, schops2019bad, sucar2021imap, sang2023high}. 
In mobile devices, depth sensors are often not available, restricted in range, or not delivering the desired quality, thus we desire techniques that can recover 3D structure from image streams. Here, our focus is on the recovery of 3D structure specifically from monocular images.
DTAM~\cite{newcombe2011dtam}, SVO~\cite{forster2014svo}, REMODE~\cite{pizzoli2014remode}, and LSD SLAM~\cite{engel2014lsd}  recover the dense or semi-dense depth map by photometric alignment of motion-stereo images. SVO~\cite{forster2014svo}, REMODE~\cite{pizzoli2014remode} and the work proposed by Mur-Artal et al.~\cite{mur2015probabilistic}  utilize image intensity observations from sequential frames with sufficient parallax to probabilistically refine depth estimates and reject outliers effectively. The uncertainty of estimated depth values for corresponding pixels can be iteratively reduced with sequential observations.

The integration of IMU measurements can enhance the accuracy and robustness of monocular dense mapping by providing scale information.
To address the perception needs of aerial robots, Yang et al.~\cite{yang2017real} propose a method for generating a 3D mesh, which sequentially fuses local depth measurements obtained from semi-global regularization into a global map.
Sch{\"o}ps  et al.~\cite{schops2017large} employ plane-sweep-based dense depth recovery from temporal motion-stereo images, with the poses being estimated by a VINS system.
Kimera~\cite{rosinol2020kimera} also relies on poses estimated from VINS and conducts dense mesh reconstruction by sparse 3D points guided 2D Delaunay triangulation.
%


\PAR{MVS Neural Networks.} 
The MVS technique aims to recover dense depth information of a given reference image by utilizing neighboring images with known camera intrinsics and 6 degrees of freedom (DOF) relative poses. In recent years, learning-based MVS approaches have gained popularity in the 3D computer vision research field, as they can recover more accurate depth maps by leveraging learned priors not available to traditional MVS methods. 
Most of the existing MVS neural networks follow a two-stage paradigm~\cite{yao2018mvsnet, huang2018DeepMVS, im2019dpsnet,wang2018mvdepthnet, hou2019multi,gu2020cascade,yang2022non,sayed2022simplerecon}.  Firstly, a plane-sweep cost volume generation is performed by matching the image intensity or extracted (deep) features of the reference (target) image and source images. Secondly, the dense depth of the reference image is predicted by utilizing cost volume regularization modules with 2D or 3D convolutional neural networks.

Numerous MVS neural networks relying on 3D CNN regularization, like MVSNet~\cite{yao2018mvsnet}, DPSNet~\cite{im2019dpsnet}, RecurrentMVSNet~\cite{yao2019recurrent}, P-MVSNet~\cite{luo2019p}, CascadeNet~\cite{gu2020cascade}, ESTDepth~\cite{long2021multi}, and NP-CVP-MVSNet~\cite{yang2022non}, demonstrate promising performance in generating high-quality dense reconstructions, while often requiring significant computational resources.
%
Instead of relying on computationally intensive 3D convolutions, many research works including DeepMVS~\cite{huang2018DeepMVS}, MVDepthNet~\cite{wang2018mvdepthnet}, and GPMVS~\cite{hou2019multi} have formulated cost volumes by directly matching pixel intensities and utilizing 2D convolutions for regularization. Similarly, DeepVideoMVS~\cite{duzceker2021deepvideomvs} and SimpleRecon~\cite{sayed2022simplerecon} also employ 2D convolutions for regularization, but generate cost volumes from deep feature matching. These works that use 2D convolutions for regularization are more suitable for real-time applications than their 3D convolution counterparts.

Some existing MVS methods adopt a two-step approach to recover a dense depth map, where they first recover the sparse depth of certain points, and then densify the sparse depth.
Fast-MVSNet~\cite{yu2020fast} proposes a method that formulates a sparse 3D cost volume to recover sparse depth. This is followed by applying a single-view 2D CNN, which densifies the initial sparse depth by the high-resolution reference RGB image. To improve the accuracy of the predicted dense depth map of the reference image, differentiable Gauss-Newton optimization to enforce deep feature consistency with neighboring images is performed. On the contrary, Our SPA-MVSNet does not rely on a customized sparse cost volume or additional non-linear optimization. 
Poggi et al.\cite{poggi2022guided} present a method to guide MVS depth prediction with sparse depth by enforcing a minimum in the variance-based cost volume near the known sparse depth. Similarly, the work proposed by Qi et al. \cite{qi2022sparse} modulates the cost volume according to depth priors to minimize the variance-based cost volume near sparse depth priors, and then utilizes 3D deformable CNN for cost volume regularization. In contrast to these two works, our SPA-MVSNet utilizes a dense sparse depth prior to narrow the 3D search space of the cost volume, instead of directly modulating the cost volume with sparse depth.
%

%
In addition to the cost-volume based MVS techniques mentioned above, a new category of learning-based volumetric methods has recently emerged, which directly extract 3D scene structures from deep feature volumes~\cite{murez2020atlas, sun2021neuralrecon, bozic2021transformerfusion, stier2021vortx}. This stream of works avoids fusing multiple dense depth maps incrementally to create a global map. Instead, they enable to learn global shape priors and surface smoothness from data, resulting in a globally coherent 3D scene reconstruction.
%

\PAR{Learning-based Dense SLAM.}
CNN-SLAM~\cite{tateno2017cnn} is considered as one of the pioneering works that utilizes single-view depth prediction from CNN neural networks for dense mapping.
However, optimizing depth values of a high-dimensional dense depth map can be computationally intensive. To address this, CodeSLAM~\cite{bloesch2018codeslam} introduces a method for efficient dense depth optimization using a CVAE network architecture that predicts and encodes dense depth from a single RGB image. The low-dimensional depth code can be effectively optimized in a state estimator.
CodeVIO~\cite{zuo2021codevio} further incorporates depth of sparse points from a VIO system into the CVAE for dense depth completion and encoding. 
%
%
CodeMapping~\cite{matsuki2021codemapping} integrates depth code optimization into a full SLAM system ~\cite{ORBSLAM3_TRO}, and reconstructs a local dense map in a separate thread.

Mobile3DRecon~\cite{yang2020mobile3drecon} employs a multi-view semi-global matching method to recover a dense depth map, which is subsequently refined by a lightweight CNN-based single-view depth refinement neural network.
DroidSLAM~\cite{teed2021droid}, a fully differentiable learned dense SLAM system, achieves high accuracy in pose estimation and remarkable generalization capability through accurate dense optical prediction~\cite{teed2020raft} and differentiable dense bundle adjustment. However, it does not focus on 3D dense mapping.
GeoRefine~\cite{ji2022georefine} integrates geometric SLAM based on dense optical flow measurements with a single-view depth prediction model. The depth values in the dense depth map are then optimized with multiple types of geometric constraints to recover coherent 3D geometry over the image sequence.

In the realm of related research, our proposed method aligns most closely with TANDEM~\cite{koestler2021tandem}.  It is a real-time monocular dense mapping system that obtains poses from Direct Sparse Odometry~\cite{engel2017direct} and incrementally fuses MVS-predicted dense depth maps by TSDF fusion~\cite{whelan2012kintinuous}. In comparison to TANDEM~\cite{koestler2021tandem}, our proposed method offers several key advantages. Firstly, we leverage the noisy sparse depth cues from VIO landmarks to guide cost volume generation in our efficient SPA-MVSNet, which improves both the accuracy and generalization of MVS depth prediction. 
TANDEM~\cite{koestler2021tandem} employs an MVS neural network with 3D CNN cost volume regularization, whereas our SPA-MVSNet relies on a lighter-weight 2D CNN counterpart, resulting in a significant improvement in efficiency.
Secondly, compared to the estimated up-to-scale trajectory and dense reconstruction from monocular images in TANDEM~\cite{koestler2021tandem}, our system, with a visual-inertial setup, enables accurate poses and dense mapping in metric scale.
Thirdly, our method is integrated into a feature-based VIO system~\cite{ORBSLAM3_TRO}, which provides more robust and accurate pose estimation compared to the direct photometric visual odometry. Consequently, our method allows for high-quality dense mapping under challenging scenarios with fast motion. 

\label{sec:method}
\section{System Overview}
\begin{figure*}[ht!]
	\centering
	\includegraphics[width=0.8\textwidth]{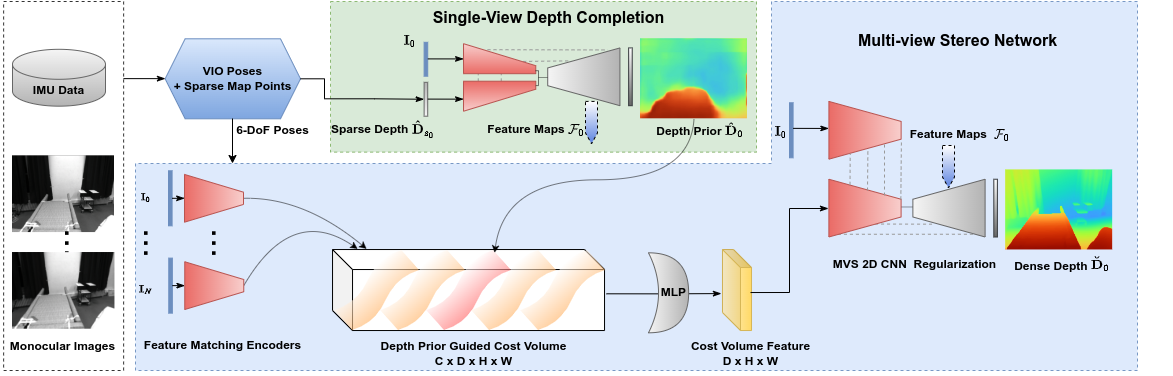}
	\vspace{-1em}
	\caption{System overview of SimpleMapping. VIO takes input of monocular images and IMU data to estimate 6-DoF camera poses and generate a local map containing noisy 3D sparse points. Dense mapping process first performs the single-view depth completion with VIO sparse depth map $\hat{\mathbf{D}_{s_0}}$ and the reference image frame $\mathbf{I}_0$, then adopts multi-view stereo (MVS) network to infer a high-quality dense depth map for $\mathbf{I}_0$. The depth prior $\hat{\mathbf{D}}_0$ and hierarchical deep feature maps $\mathbf{\mathcal{F}}_0$  from the single-view depth completion contribute to the cost volume formulation and 2D CNN cost volume regularization in the MVS. The high-quality dense depth prediction from the MVS, $\breve{\mathbf{D}}_0$, is then fused into a global TSDF grid map for a coherent dense reconstruction.}
	\label{fig:system_overview}
	\vspace{-1em}
\end{figure*}

The proposed real-time dense mapping system is depicted in Figure~\ref{fig:system_overview}. It comprises two main processes: VIO pose tracking (ORB-SLAM~3~\cite{ORBSLAM3_TRO}) and Dense Mapping. The VIO uses monocular images and inertial measurements to achieve 6-DoF pose tracking and generates a local map with 3D sparse points. Additionally, it performs covisibility-based keyframe selection and generates 2D sparse depth maps by projecting 3D sparse points into the image planes.
In the dense mapping process, the selected keyframe images and 2D sparse depth map are input to our proposed sparse point aided multi-view stereo network (SPA-MVSNet) to recover dense depth maps. To achieve a globally consistent 3D dense reconstruction, the dense depth maps are incrementally fused into a TSDF volumetric map, which is managed by voxel-hashing~\cite{niessner2013real}. The 3D dense mesh can be recovered from the TSDF volume by Marching Cubes algorithm~\cite{lorensen1987marching}. Our system seamlessly integrates the VIO pose tracking and Dense Mapping processes and executes them in parallel threads, enabling real-time tracking and high-quality dense mapping in a coupled way.

\subsection{Sparse Visual-Inertial Odometry}
\label{sec:sparseVIO}
The feature-based VIO systems have reached a relatively advanced stage of development ~\cite{mourikis2007multi,leutenegger2015keyframe,qin2018vins,ORBSLAM3_TRO,lang2022ctrl} to deliver robust 6-DoF pose estimations. In our dense mapping system, ORB-SLAM3~\cite{ORBSLAM3_TRO}, renowned for its superior performance, is chosen for pose tracking with loop closure disabled. Moreover, we adapt it to carefully select a set of MVS keyframes and generate 2D sparse depth maps of reference keyframes, which are essential for the dense mapping process.


\PAR{Keyframe Selection.} Selecting dedicated MVS keyframes from the available pose tracking keyframes in ORB-SLAM3~\cite{ORBSLAM3_TRO} is required. We feed a window of eight keyframes as input to the MVS. To ensure a balance between triangulation quality and view frustum overlap, we adopt the co-visible connections and pose distance criteria suggested in \cite{duzceker2021deepvideomvs, hou2019multi}. This approach attempts to cover as much space as possible while preserving the affinity of the reference keyframes. Co-visibility, defined as the number of common landmarks observed by both frames, determines the strength of the relationship between pose tracking keyframes and their overall importance in the scene structure. We eliminate co-visible keyframes with few connections and those with relative pose distances ($\rm{dist}(\cdot)$) greater than a threshold value of $p_{\rm{th}}$. The filtered keyframes are then ranked based on the defined penalty metric:
\begin{equation}
	\label{eq:pose_distance}
	\begin{array}{l}
	    \rm{dist}(\mathbf{T}_{0,i}) = \sqrt{\parallel \mathbf{t}_{0,i} \parallel + \frac{2}{3}tr(\mathbb{I} - \mathbf{R}_{0,i})} \\
    	\rm{penalty}(\mathbf{T}_{0,i}) = \alpha (\parallel \mathbf{t}_{0,i} \parallel - t_{\rm{th}})^2 + \frac{2}{3}\rm{tr}(\mathbb{I} - \mathbf{R}_{0,i})\\
    	\alpha = \left\{ \begin{array}{rcl} 5.0 & \mbox{if} \parallel \mathbf{t}_{0,i} \parallel \le t_{\rm{th}} \\ 1.0 & \mbox{if} \parallel \mathbf{t}_{0,i} \parallel > t_{\rm{th}}  \end{array} \right.
	\end{array}
\end{equation}
where $\mathbf{T}_{0,i} = [\mathbf{R}_{0,i}, \mathbf{t}_{0,i}]$ denotes the relative rotation and translation between reference and source keyframes, indexed by $0$ and $i, \; i \in \{1, 2, \cdots, N\}$, respectively.
$t_{\rm{th}}$ represents the translation threshold for penalty computation, $\mathbb{I}$ denotes the identity matrix. We set $p_{\rm{th}} = 0.20$m and $t_{\rm{th}} = 0.25$m in our experiments. We select a window of eight keyframes with the lowest penalty metrics as MVS keyframes. The newest keyframe in the window serves as the reference image while the others serve as source images for the MVS network.

\PAR{Sparse Depth Filtering.} To generate a 2D sparse depth map for MVS, we project 3D sparse VIO points allocated within the reference keyframe frustum onto the image plane. Our experimental findings demonstrate that both the number and accuracy of the valid depth values in the sparse depth map can affect the quality of dense depth depth prediction. In order to avoid undesired artifacts in the depth prediction caused by extreme outliers in sparse depths, we filter out 3D sparse points based on a depth upper threshold $d_{\rm{th}}$ and reprojection errors $r_{\rm{th}}$ in the corresponding images. We select $d_{\rm{th}} = 5.0$m and $r_{\rm{th}} = 2.0$ pixels as the threshold values, resulting in the retention of $80-300$ points with valid depth values in the projected 2D depth maps -- while all other pixels lacking sparse depth are assigned zero values.

\subsection{Sparse Points-Aided Single-View Depth Completion}
~\label{sec:depthcompletion}

We propose SPA-MVSNet, which effectively utilizes the sparse depth points obtained from VIO for accurate dense depth prediction.
%
%
Prior to the MVS depth prediction, we employ a lightweight CNN to perform single-view depth completion on a reference image and its corresponding 2D sparse depth map generated using the method described in Section ~\ref{sec:sparseVIO}. This results in a complete single-view dense depth map -- which is then used as a rough geometric prior to guide the cost volume construction and regularization in the MVS process (as described in Section~\ref{sec:multi-view}). 

%

To effectively densify the sparse depth map, we employ a preprocessing step involving a sparse-to-dense module~\cite{wong2021unsupervised} comprised of a series of Min- and Max-pooling layers with various kernel sizes and multiple convolutional layers. This results in a coarse dense depth map, which is then passed through an encoder-decoder-based depth completion network, as depicted in Figure~\ref{fig:system_overview}. The single-view depth completion network, based on~\cite{wong2021unsupervised}, consists of two encoder branches, one for the RGB image and the other for the corresponding 2D sparse depth map. Both encoder branches have the same architecture, which includes multiple downsampling VGG blocks~\cite{ma2019self} with skip connections. Finally, the decoder takes the deep features from both encoders to predict a dense depth map at the last layer.

We supervise the depth completion network with ground-truth dense depth maps. The training loss is the linear combination of two terms:
\begin{equation}
	\label{eq:depth_comp_loss0}
	\mathcal{L} = {\omega}_{rec} \mathcal{L}_{rec} + {\omega}_{sm} \mathcal{L}_{sm},
\end{equation}
where $\mathcal{L}_{rec}$ denotes depth reconstruction loss and $\mathcal{L}_{sm}$ the smoothness loss. To balance the contribution of these two terms in the loss function, we have experimentally chosen equal weights for them. Specifically, for the depth reconstruction loss, we aim to minimize the $L_1$ error between the predicted depth map $\hat{\mathbf{D}}$ and the ground truth depth map ${\mathbf{D}_{gt}}$ by:
\begin{equation}
	\label{eq:depth_comp_loss1}
	\mathcal{L}_{rec} = \frac{1}{|\Omega|} \sum_{\mathbf{u} \in \Omega} \mid \hat{\mathbf{D}}(\mathbf{u}) - \mathbf{D}_{gt}(\mathbf{u}) \mid,
\end{equation}
where $\mathbf{u} \in \Omega$ denotes the valid pixel coordinates in the ground truth depth map.
%

The depth reconstruction loss considers errors on individual depth values, failing to incorporate neighborhood constraints, which often results in significant discontinuities. To address this issue, we introduce local smoothness constraints by imposing an $L_1$ penalty on depth gradients in the horizontal and vertical directions (denoted as $U$ and $V$ respectively) in the image plane. These penalties are weighted by their respective image gradients in an image $\mathbf{I}_t$, denoted by $\lambda_U(\mathbf{u}) = e^{- \mid \partial_U \mathbf{I}_t(\mathbf{u}) \mid}$ and $\lambda_V(\mathbf{u}) = e^{- \mid \partial_V \mathbf{I}_t(\mathbf{u}) \mid}$: 
\begin{equation}
	\label{eq:depth_comp_loss2}
	\scalemath{.9}{
	\mathcal{L}_{sm} \! = \!\frac{1}{|\Omega|} \sum_{\mathbf{u} \in \Omega} \lambda_U(\mathbf{u}) \mid \partial_U \hat{\mathbf{D}}(\mathbf{u}) \mid \!+\! \lambda_V(\mathbf{u}) \mid \partial_V \hat{\mathbf{D}}(\mathbf{u}) \mid.
	}
\end{equation} 
%
%

\subsection{Multi-View Stereo with Dense Depth Prior}
\label{sec:multi-view}
We guide the MVS depth prediction by the dense depth prior generated by the single-view depth completion network described above. 
Our MVS network is built on the foundation of SimpleRecon~\cite{sayed2022simplerecon}, a highly efficient network that uses 2D CNNs instead of more computationally demanding 3D counterparts. 
%
%


\PAR{Cost Volume Guided by Depth Priors.}
In SimpleRecon~\cite{sayed2022simplerecon}, a 4D cost volume with dimensions $C \times D \times H \times W$  is constructed from $D$ fronto-parallel hypothesis planes that are evenly distributed at varying distances from the camera center. $ H, \; W$ of the cost volume denote the height and width of the input images, respectively.
The cost volume is filled with carefully computed keyframe and geometric metadata of dimensions $C$, which includes dot products of deep features extracted from images, ray directions, reference plane depth, and reference frame projected depth, etc. Although simple, this approach has been demonstrated to greatly benefit the final dense depth prediction.
The original SimpleRecon~\cite{sayed2022simplerecon} constructs the cost volume covering the 3D space from 0.25m to 5m, with 64 depth planes with a distance interval of 0.0754m. However, it is important to note that most of the cost volume is empty and physical surfaces are only allocated in a small portion of the 4D cost volume.
To reduce the excessive computation on the vast empty space in the cost volume and focus on regions more likely to yield surfaces for precise depth recovery, we carefully direct the cost volume generation using the rough dense depth prior from single-view depth prediction, denoted as $\hat{\mathbf{D}}$. This depth prior is defined as a center surface, and we allocate other depth hypotheses surfaces at consistent intervals on both sides of the center surface. In the cost volume, the hypothesis 3D surfaces are spaced at distances:
\begin{align}
	\label{eq:cost_volume}
	\mathcal{C}(\mathbf{u}) = \{ & \hat{\mathbf{D}}(\mathbf{u}) - n_1 \lambda , \hat{\mathbf{D}}(\mathbf{u}) - (n_1 - 1)\lambda,   \cdots, \hat{\mathbf{D}}, \\ \nonumber
	& \cdots, \hat{\mathbf{D}}(\mathbf{u}) + (n_2 -1)\lambda , \hat{\mathbf{D}}(\mathbf{u}) + n_2 \lambda \},
\end{align}
where $\lambda$ denotes the constant interval, $n_1$ and $n_2$ the number of hypothesis surfaces on both sides of the central prior surface.
In our implementation, we generate 64 evenly distributed hypothesis surfaces with a constant interval of 0.04m (\ie, $\lambda=0.04$, $n_1=31$ and $n_2=32$). Guided by depth prior, this approach significantly reduces the size of the hypothesis space used for generating the cost volume, leading to more accurate dense depth predictions.

Meta data in 4D cost volume with dimension $C \times D \times H \times W$ is further processed by a multilayer perceptron (MLP), which condenses the cost volume into a feature map with dimension of $1 \times D \times H \times W$. Following a similar approach to SimpleRecon~\cite{sayed2022simplerecon}, the condensed feature map is then passed through a 2D CNN-based MVS encoder-decoder pipeline, which predicts the final dense depth map $\breve{\mathbf{D}}$.


\PAR{Deep Depth Features Fused into MVS Encoder-Decoder.}
%
The deep features extracted from the single-view depth completion network  (see Section~\ref{sec:depthcompletion}) capture essential depth prior information, including details of sparse depth inputs and geometric interrelationships of pixels within the reference image. To ensure the cost volume regularization module in the MVS is aware of the depth prior that guides the cost volume generation, hierarchical deep feature maps from the depth completion decoder are concatenated into the decoder of the MVS regularization module through skip connections.
The concatenated features are then fused and refined in the MVS decoder at multiple scales, leading to a notable improvement in the accuracy of the final dense depth prediction.

\subsection{Network Training}~\label{sec:network training}

\PAR{Training Dataset.} Our proposed network, SPA-MVSNet, is trained on the ScanNet~\cite{dai2017scannet}, which comprises indoor RGB-D images captured using a Kinect RGB-D camera. To train the model on this RGB-D dataset, additional to provided image and depth data, camera poses and intrinsics, we also require sparse depths that are similar to the noisy 3D sparse points estimated in a VIO system. To generate the sparse depth map input, we employ FAST~\cite{rosten2006machine} keypoint detection on an RGB image and select the 250 keypoints with the highest scores. The depth values of these selected keypoints are extracted from the provided ground truth depth maps. To simulate the 3D sparse points in a running VIO accurately, we add Gaussian noise to the sampled sparse points from the depth map. The noise is applied to the intrinsics, camera poses, and depth values, accounting for errors in camera calibration (focal length $f_{\rm{x}}, f_{\rm{y}}$, and principal point offset $c_{\rm{x}}, c_{\rm{y}}$), camera poses (orientation $\mathbf{R}$ and translation $\mathbf{t}$), and the triangulation process of 3D sparse points (depth value $d$ and projected pixel coordinates $u, v$). Our default noise configuration is $\sigma_{f_{\rm{x}}} = \sigma_{f_{\rm{y}}} = 0.1$mm, $\sigma_{c_{\rm{x}}} = \sigma_{c_{\rm{y}}} = 0.1$ pixels,  $\sigma_{{R}} =0.01$\textdegree, $\sigma_{{t}} = 0.005$m, $\sigma_d = 0.45$m, $\sigma_{uv} = 3.0$ pixels.


\PAR{Training procedure.}  For the MVS training and testing, approximately 0.4 million keyframe images are selected from the ScanNet~\cite{dai2017scannet}. The frame selection strategy follows the approach of ~\cite{duzceker2021deepvideomvs, sayed2022simplerecon}, and the official training and test split is used. The SPA-MVSNet is trained in three stages. Firstly, the single-view depth completion network is trained for 8 epochs to achieve optimal performance. Secondly, the MVS network is trained using a window of posed keyframes and the dense depth map generated by the trained single-view depth completion network. Lastly, both the single-view depth completion network and the MVS network are fine-tuned together.

\label{sec:experiments}
\section{Experimental Results}

Our experiments consist of two main parts. In the first part, we evaluate the effectiveness of our SPA-MVSNet using given ground-truth camera poses and simulated noisy sparse points. Specifically, we evaluate the quality of MVS dense depth prediction and 3D reconstruction on the ScanNet~\cite{dai2017scannet} and 7-Scenes ~\cite{shotton2013scene} datasets. To evaluate 3D reconstruction, we apply the standard volumetric TSDF fusion technique to fuse different depth maps. The generalization ability of SPA-MVSNet is reflected in its performance on the unseen 7-Scenes~\cite{shotton2013scene} dataset.

In the second part, we evaluate the efficacy of SimpleMapping as a real-time dense mapping system. 
We conduct a thorough evaluation of dense depth estimation, 3D reconstruction quality, and run-time efficiency on widely used SLAM datasets such as EuRoC~\cite{Burri25012016} and ETH3D~\cite{schops2019bad}, as well as our self-collected dataset.
Please also refer to our \href{https://youtu.be/qn35joS5zZM} {\textbf{supplementary video}}, showcasing the proposed dense mapping system operating in real-time and the resulting reconstructed 3D meshes in various scenarios.

We use the metrics defined in \cite{eigen2014depth} to evaluate 2D dense depth. For the quality of 3D reconstruction, we employ the metrics of accuracy, completeness, chamfer $L_1$ distance, precision, recall, and F-score presented in \cite{bozic2021transformerfusion}. The distance threshold for precision, recall, and F-score is set to $5$cm.

\begin{table}[htb!]
	\caption{Depth evaluation of MVSNet. For each metrics, the best-performing method is marked in bold.}
    \vspace{-1em}
	\begin{center}
		\vspace{-1em}
		\setlength{\tabcolsep}{0.4em} 
		\resizebox{0.5\textwidth}{!}{  
			\begin{tabular}{lcccc|cccc}
				\toprule
				& \multicolumn{4}{c}{\textbf{ScanNet~\cite{dai2017scannet}}} & \multicolumn{4}{c}{\textbf{7Scenes~\cite{shotton2013scene}}} \\
				\cmidrule(lr){2-5}\cmidrule(lr){6-9}
				& Abs Diff [m]$\downarrow$  & Sq Rel [\%]$\downarrow$ & $\delta < 1.05 [\%]\uparrow$ &  $\delta < 1.25 [\%]\uparrow$ & Abs Diff [m]$\downarrow$  & Sq Rel [\%]$\downarrow$ & $\delta < 1.05 [\%]\uparrow$ &  $\delta < 1.25 [\%]\uparrow$ \\
				\midrule
				DPSNet ~\cite{im2019dpsnet} & 0.1552 &  0.0299 & 49.36 & 93.27 & 0.1966 & 0.0550 & 38.81 & 87.07\\
				MVDepthNet ~\cite{wang2018mvdepthnet}  & 0.1648 &  0.0343 & 46.71 & 92.77 & 0.2009 & 0.0623 & 38.81 & 87.70\\
				DELTAS~\cite{sinha2020deltas} & 0.1497 &  0.0276 & 48.64 & 93.78 & 0.1915 & 0.0490 & 36.36 & 88.13\\
				GPMVS ~\cite{hou2019multi}  & 0.1494 &  0.0292 & 51.04 & 93.96 & 0.1739 &  0.0462 & 42.71 & 90.32\\
				DeepVideoMVS~\cite{duzceker2021deepvideomvs} &
				 0.1186 &    0.0190 &  60.20 &  96.76 &  0.1448 &   0.0335 &   47.96 &  93.79\\
				SimpleRecon~\cite{sayed2022simplerecon} &  0.0893 &    0.0127 &   72.84 &   98.05 &  0.1045 & 0.0153  &  59.78 &  97.38\\
				\textbf{Ours} &  \textbf{0.0696} &    \textbf{0.0088} &   \textbf{79.60} &   \textbf{98.53} &  \textbf{0.0863} &    \textbf{0.0115}  &  \textbf{67.96} &  \textbf{97.93}\\
				\bottomrule
			\end{tabular}
		}
	\end{center}
	\vspace{-2em}
	\label{table:depth_scannet_results}
\end{table}

\begin{table}[htb!]
	\caption{Mesh evaluation on ScanNet~\cite{dai2017scannet} of MVSNet. The vol. column denotes whether a method is a volumetric 3D reconstruction method; depth maps estimated by other MVS methods are integrated into standard TSDF fusion for reconstruction.}
\vspace{-1em}
	\begin{center}
		\vspace{-1em}
		\setlength{\tabcolsep}{0.1em} 
		\resizebox{\columnwidth}{!}{  
			\footnotesize
			\begin{tabular}{lcccccccc}
				\toprule
				& Vol. & Comp [cm]$\downarrow$  & Acc [cm]$\downarrow$ & Chamfer $L_1$ [cm]$\downarrow$ & Prec [\%]$\uparrow$ & Recall [\%] $\uparrow$ & F-Score [\%]$\uparrow$\\
				\midrule
				MVDepthNet~\cite{wang2018mvdepthnet}   & No & 12.94 & 8.34 & 10.64 & 44.3 & 48.7 & 46.0  \\
				ESTDepth~\cite{long2021multi} & No & 12.71 & 7.54 & 10.12 & 45.6 & 54.2 & 49.1  \\
				DELTAS~\cite{sinha2020deltas} & No & 11.95 & 7.46 & 9.71 & 47.8 & 53.3 & 50.1  \\
				DeepVideoMVS~\cite{duzceker2021deepvideomvs} & No & 10.68 & 6.90 & 8.79 & 54.1 & 59.2 & 56.3  \\
				COLMAP~\cite{schoenberger2016mvs} & No & 10.22 & 11.88 & 11.05 & 50.9 & 47.4 & 48.9 \\
				ATLAS~\cite{murez2020atlas} & Yes & 7.16 & 7.61 & 7.38 & 67.5 & 60.5 & 63.6 \\
				NeuralRecon~\cite{sun2021neuralrecon} & Yes &  5.09 & 9.13 & 7.11 & 63.0 & 61.2 & 61.9  \\
				3DVNet~\cite{rich20213dvnet} & Yes & 7.72 &  6.73 & 7.22 & 65.5 & 59.6 & 62.1 \\
				TransformerFusion~\cite{bozic2021transformerfusion} & Yes &  5.52 & 8.27 & 6.89 &  72.8 & 60.0 & 65.5\\
				VoRTX~\cite{stier2021vortx} & Yes &  \textbf{4.31} & 7.23 &  5.77 &  \textbf{76.7} &  65.1 &  70.3  \\
				SimpleRecon~\cite{sayed2022simplerecon} & No & 5.53 &  6.09 &  5.81 & 68.6 &  65.8 &  67.1 \\
				\textbf{Ours} & No & 5.23 &  \textbf{4.57} &  \textbf{4.90} &  75.7 &  \textbf{72.7} &  \textbf{74.0} \\
				\bottomrule
			\end{tabular}
		}
	\end{center}
	\label{table:MVSmesh_results}
	\vspace{-2em}
\end{table}

\subsection{Dense Depth Prediction of SPA-MVSNet}
\label{sec:experiments_mvsnet}


In Table~\ref{table:depth_scannet_results}, we present a comparison of our method with state-of-the-art MVS methods on both ScanNet~\cite{dai2017scannet} and 7-Scenes~\cite{shotton2013scene} in terms of depth prediction. Our SPA-MVSNet outperforms the compared methods in all metrics, demonstrating its effectiveness. Despite the sparse and noisy nature of the sampled sparse depths, which consist of only 250 points for reference keyframes in our method, they still serve as useful priors to guide network optimization, indicating the robustness of SPA-MVSNet in dealing with various types of noise.

For 3D mesh reconstruction evaluation, we divide previous methods into two groups: learning-based volumetric 3D reconstruction methods, and MVS methods that first predict depth and then fuse it into a TSDF for the final 3D reconstruction of the scene. We follow the evaluation protocol in \cite{bozic2021transformerfusion}, and present the results in Table \ref{table:MVSmesh_results}. Our method outperforms all comparisons by a large margin on the ScanNet~\cite{dai2017scannet}. Moreover, our lightweight MVS network achieves superior performance compared to volumetric fusion methods, such as TransformerFusion~\cite{bozic2021transformerfusion} and VoRTX~\cite{stier2021vortx}, which use computationally demanding self-attention mechanisms and fuse deep features from all the sequential images. These findings highlight the potential of our approach for efficient and effective 3D reconstruction.

\begin{figure}[htb!]
	\centering
	\includegraphics[width=0.5\textwidth]{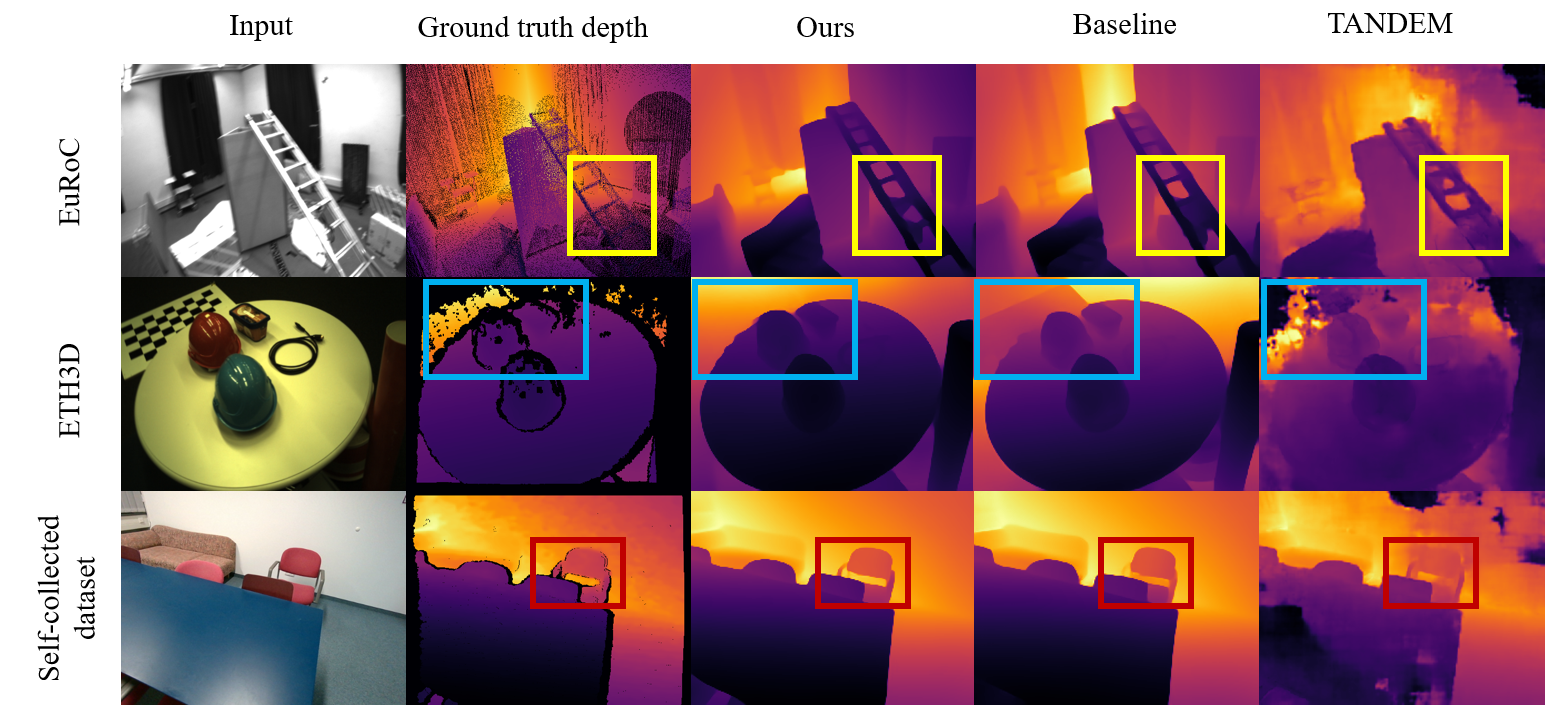}
	\vspace{-2em}
	\caption{Depth prediction from the MVS networks constituting dense mapping systems on EuRoC~\cite{Burri25012016}, ETH3D~\cite{schops2019bad} and our self-collected dataset. 
	The highlighted boxes demonstrate finer details of depth prediction by our method compared to others.}
	\label{fig:depth_all}
	\vspace{-1em}
\end{figure}

\subsection{Real-time Dense Mapping}


We evaluate our complete dense mapping system against other monocular mapping systems, \ie DeepFactors~\cite{czarnowski2020deepfactors}, TANDEM~\cite{koestler2021tandem} and Sigma-Fusion~\cite{rosinol2022probabilistic}, as well as visual-inertial real-time systems, CodeMapping~\cite{matsuki2021codemapping} and Kimera~\cite{rosinol2020kimera}. In addition, we aim to demonstrate the efficacy of incorporating noisy sparse depth maps into SPA-MVSNet by comparing our system to a \textbf{baseline} system that uses pose tracking from ORB-SLAM3~\cite{ORBSLAM3_TRO} and dense depth prediction from SimpleRecon~\cite{sayed2022simplerecon}. 
%
It is worth noting that both CodeMapping \cite{matsuki2021codemapping} and the baseline method employ the same VIO tracking front-end, ORB-SLAM3~\cite{ORBSLAM3_TRO}, as ours.
%
%
Furthermore, we showcase the comparable reconstruction performance of SimpleMapping utilizing only a monocular camera setup without IMU, against a state-of-the-art RGB-D dense SLAM method with neural implicit representation, Vox-Fusion\cite{yang2022vox}.
All experiments conducted for different methods utilize the best model and configuration that have been released.

\definecolor{Gray}{gray}{0.9}
\begin{table}[htb!]
	\begin{center}
	 \caption{Depth evaluation on EuRoC~\cite{Burri25012016} V1 and V2 of mapping systems. 
	 Depth evaluation results of rendered dense map from generated mesh are presented in gray.} 
    \label{table:depth_euroc_results}
	\setlength{\tabcolsep}{0.4em}
	\resizebox{0.95\columnwidth}{!}{
	\begin{tabular}{clccccccc}
    	\toprule
         &   & V101   & V102   & V103   & V201   & V202   & V203   & Avg. \\
        \midrule
        {\multirow{6}{*}{AbsDiff [m]$\downarrow $}} & DeepFactors\cite{czarnowski2020deepfactors}  & 0.842  & 0.875  & 0.833  & 0.859  & 1.29   & 1.08   & 0.963   \\ 
        & CodeMapping\cite{matsuki2021codemapping} & \textbf{0.192}  & 0.259  & 0.283  & 0.290  & 0.415  & 0.686  & 0.354   \\ 
        & TANDEM\cite{koestler2021tandem}      & 0.283 & 0.226 & 0.377 & \textbf{0.221} & 0.285 & 0.688 & 0.347  \\  
        & Baseline    & 0.309  & 0.286  & 0.285  & 0.293  & 0.386  & 0.246  & 0.301   \\ 
        & \textbf{Ours}        & 0.251  & \textbf{0.214}  & \textbf{0.262}  & 0.246  & \textbf{0.316}  & \textbf{0.189}  & \textbf{0.246}  \\
        & \cellcolor{Gray}\textbf{Ours (fused)}        &   \cellcolor{Gray}0.221 & \cellcolor{Gray}0.157  & \cellcolor{Gray}0.188  & \cellcolor{Gray}0.162 & \cellcolor{Gray}0.192  & \cellcolor{Gray}0.114  & \cellcolor{Gray}0.172   \\
        \midrule
        {\multirow{6}{*}{RMSE [m]$\downarrow $}}    & DeepFactors\cite{czarnowski2020deepfactors}  & 1.05   & 1.03   & 0.940  & 1.02   & 1.53   & 1.25   & 1.14    \\  
        & CodeMapping\cite{matsuki2021codemapping} & \textbf{0.381}  & 0.369  & 0.407  & 0.428  & 0.655  & 0.952  & 0.532   \\ 
        & TANDEM\cite{koestler2021tandem}      & 0.457 & 0.366 & 0.524 & \textbf{0.379} & 0.513 & 0.786 & 0.504  \\ 
        & Baseline    & 0.477  & 0.407  & 0.390  & 0.440  & 0.555  & 0.354  & 0.437   \\  
        & \textbf{Ours}        & 0.411  & \textbf{0.334}  & \textbf{0.374}  & 0.385  & \textbf{0.492}  & \textbf{0.287}  & \textbf{0.380}   \\ 
        & \cellcolor{Gray}\textbf{Ours (fused)}        & \cellcolor{Gray}0.377  & \cellcolor{Gray}0.280  & \cellcolor{Gray}0.296  & \cellcolor{Gray}0.318  & \cellcolor{Gray}0.383  & \cellcolor{Gray}0.216  & \cellcolor{Gray}0.321   \\ 
        \midrule
        {\multirow{6}{*}{$\delta < 1.25 [\%]\uparrow $}}   & DeepFactors\cite{czarnowski2020deepfactors}  & -      & -      & -      & -      & -      & -      & -       \\ 
        & CodeMapping\cite{matsuki2021codemapping} & -      & -      & -      & -      & -      & -      & -       \\  
        & TANDEM\cite{koestler2021tandem}      & 88.57  & 92.17  & 82.47  & {88.86}  & {88.01}  & 57.46  & 82.92   \\ 
        & Baseline    & 87.75  & 90.00  & 91.19  & 88.28  & 85.51  & 91.62  & 89.06   \\  
        & \textbf{Ours}        & \textbf{91.11}  & \textbf{93.17}  & \textbf{92.61}  & \textbf{90.79}  & \textbf{88.46}  & \textbf{94.68}  & \textbf{91.80}   \\ 
        & \cellcolor{Gray}\textbf{Ours (fused)}        & \cellcolor{Gray}92.53  & \cellcolor{Gray}94.83  & \cellcolor{Gray}95.80  & \cellcolor{Gray}94.06 & \cellcolor{Gray}93.74  & \cellcolor{Gray}97.33  & \cellcolor{Gray}94.44   \\
        \bottomrule
    \end{tabular}
    }
    \end{center}
    \vspace{-2em}
\end{table}
	
\begin{table}[htb!]
    \caption{Depth evaluation on ETH3D~\cite{schops2019bad} and self-collected dataset of mapping systems.}
    \vspace{-1em}
	\begin{center}
	    \vspace{-1em}
		\setlength{\tabcolsep}{0.4em} 
		\resizebox{0.5\textwidth}{!}{  
			\begin{tabular}{clcccc|ccccc}
				\toprule
				& & \multicolumn{4}{c}{\textbf{ETH3D}} & \multicolumn{5}{c}{\textbf{Self-collected Dataset}} \\
				\cmidrule(lr){3-6}\cmidrule(lr){7-11}
				& & table\_3 & table\_4 & table\_7 & Avg. & Room 1 & Room 2 & Room 3 & Room 4 & Avg. \\
				\midrule
				{\multirow{3}{*}{SqRel [\%]$\downarrow $}} & TANDEM\cite{koestler2021tandem}   & 0.0117   & 0.0124   & 0.0374 & 0.0205  & 0.0644 & 0.0743  & 0.0245  & 0.0411  & 0.0511  \\
				  & Baseline & 0.0103   & 0.0108   & 0.0250   & 0.0153  & 0.0223 &  0.0174      &0.0748                & 0.0185     & 0.0333        \\                            
				  & Ours     & \textbf{0.0080}   & \textbf{0.0079}   & \textbf{0.0088} & \textbf{0.0082}  & \textbf{0.0140} & \textbf{0.0134} & \textbf{0.0147}    & \textbf{0.0154}  & \textbf{0.0144}\\
				\midrule
				{\multirow{3}{*}{RMSE [m]$\downarrow $}}   & TANDEM\cite{koestler2021tandem}   & 0.1203   & 0.1176   & 0.1464   & 0.1281  & 0.2845   & 0.3324         & 0.1872         & 0.2293         & 0.2584  \\ 
                & Baseline & 0.1296   & 0.1303   & 0.1661    & 0.1420  & 0.1721  &    0.1709   & 0.2670 &  0.1587  & 0.1922\\ 
                & Ours     & \textbf{0.1155}   & \textbf{0.1153}   & \textbf{0.1088}    & \textbf{0.1132}  & \textbf{0.1539}  &         \textbf{0.1510}   & \textbf{0.1506}  & \textbf{0.1426}   & \textbf{0.1495}  \\ 
				\midrule
				{\multirow{3}{*}{$\delta <1.25 [\%]\uparrow$}} & TANDEM\cite{koestler2021tandem}   & 97.85    & 97.79    & 90.16     & 95.27  & 87.48          & 85.38          & 94.79          & 90.82          & 89.62   \\ 
                & Baseline & 98.93    & 98.79    & 76.76     & 91.49   &  95.27  & 98.44  &  85.51  & 95.96               &  93.80       \\ 
                & Ours     & \textbf{99.15}    & \textbf{99.18}    & \textbf{98.55}     & \textbf{98.96}   & \textbf{98.38}          & \textbf{98.85}          & \textbf{98.21}          & \textbf{96.05}          & \textbf{97.87}   \\ 
				\bottomrule
			\end{tabular}
		}
	\end{center}
	\vspace{-2em}
	\label{table:depth_ethself_results}
\end{table}

\PAR{2D Dense Depth Evaluations.} In order to evaluate our method on the EuRoC dataset~\cite{Burri25012016}, which lacks ground truth depth maps, we generate reference depth maps by projecting the provided ground truth point cloud onto the image planes. However, as the ground truth point cloud is not densely populated, physically occluded points may be projected onto the image planes during depth map rendering, resulting in errors. To mitigate this issue, we only consider depth errors within $3$m for all methods evaluated.

Table \ref{table:depth_euroc_results} presents results on all the V1 and V2 sequences of the EuRoC~\cite{Burri25012016}, including published results from DeepFactors~\cite{czarnowski2020deepfactors} and CodeMapping~\cite{matsuki2021codemapping}, as well as our running results for TANDEM~\cite{koestler2021tandem}. 
To ensure the reliability of the evaluations, we conduct three tests on each sequence and report the averaged results. As shown in Table~\ref{table:depth_euroc_results}, our proposed approach outperforms TANDEM~\cite{koestler2021tandem} by a significant margin, exhibiting superior performance particularly on the challenging V103 and V203 sequences. 
These results demonstrate the strong generalization ability of SPA-MVSNet and the robustness of our complete dense mapping system.
In particular, depth evaluation of rendered dense depths from generated mesh are marked in gray, improving the evaluate metrics by a large margin. This indicates that online fusion makes a great contribution to fusing inaccurate depth predictions and reconstructing globally consistent dense maps.

Similar dense mapping experiments are conducted on three sequences 
from the ETH3D dataset~\cite{schops2019bad} and a self-collected dataset (comprising 4 sequences) captured by an RGB-D inertial camera. Table \ref{table:depth_ethself_results} demonstrates our method's superior performance across all metrics compared to other methods. Furthermore, comparing our method with the baseline system underscores the effectiveness of incorporating noisy sparse depths from VIO into our SPA-MVSNet.

Figure~\ref{fig:depth_all} presents qualitative results of our depth prediction on the three aforementioned datasets. Our approach achieves accurate depth predictions, even in regions lacking ground truth, while ensuring smoothness of the predicted depths. In contrast, TANDEM~\cite{koestler2021tandem} yields depth maps with high levels of noise, indicating a significant disadvantage compared to our method. Additionally, our method excels in predicting detailed structures, such as the ladder, chessboard, and chair highlighted in the figure. 
These findings convincingly demonstrate the effectiveness of our proposed method in generating high-quality depth maps even in challenging scenarios.

\begin{figure}[htb!]
	\centering
	\includegraphics[width=1.07\columnwidth]{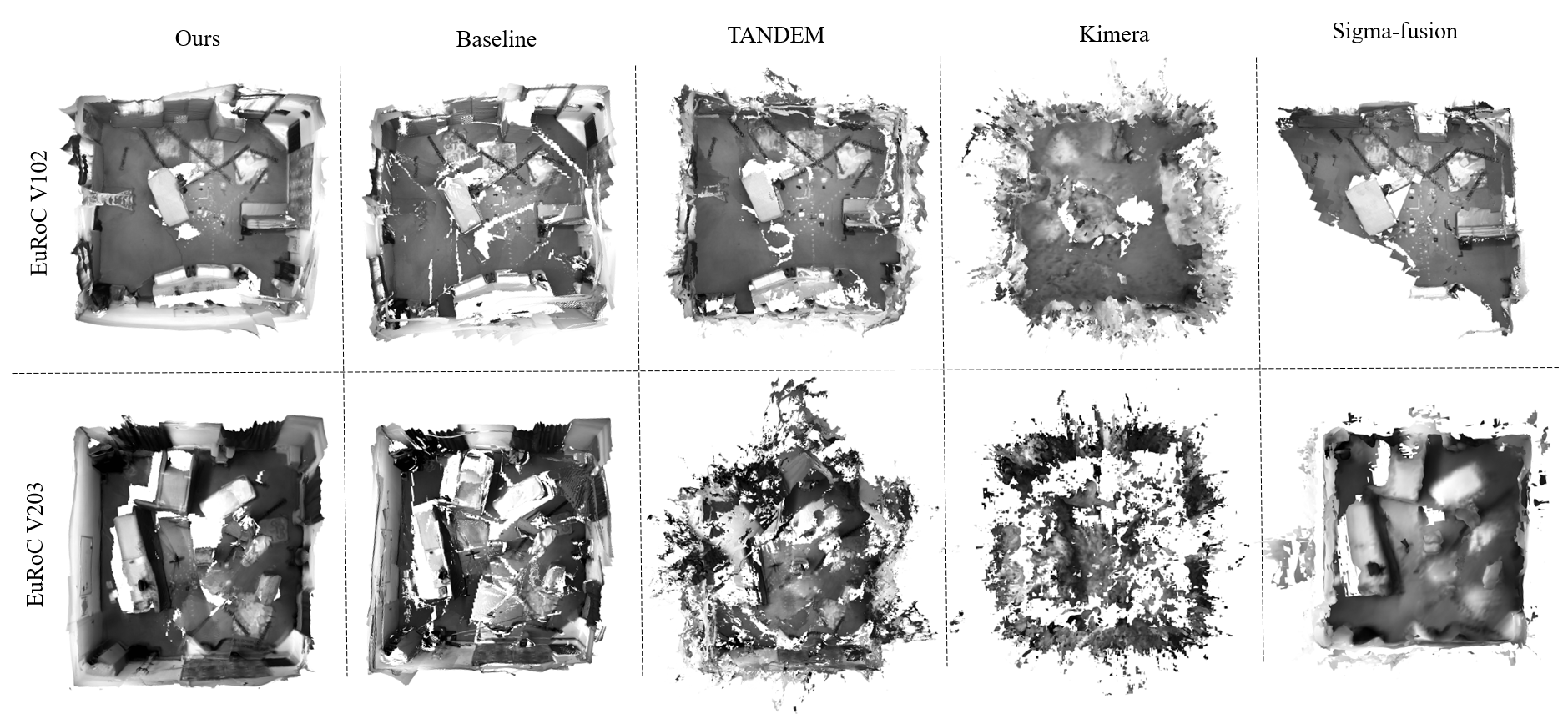}
	\vspace{-2em}
	\caption{3D reconstruction on EuRoC\cite{Burri25012016} dataset. Our method yields consistently better detailed reconstruction even in challenging scenarios.}
	\label{fig:mesh_euroc}
\end{figure}

\begin{figure}[htb!]
	\centering
	\includegraphics[width=0.9\columnwidth]{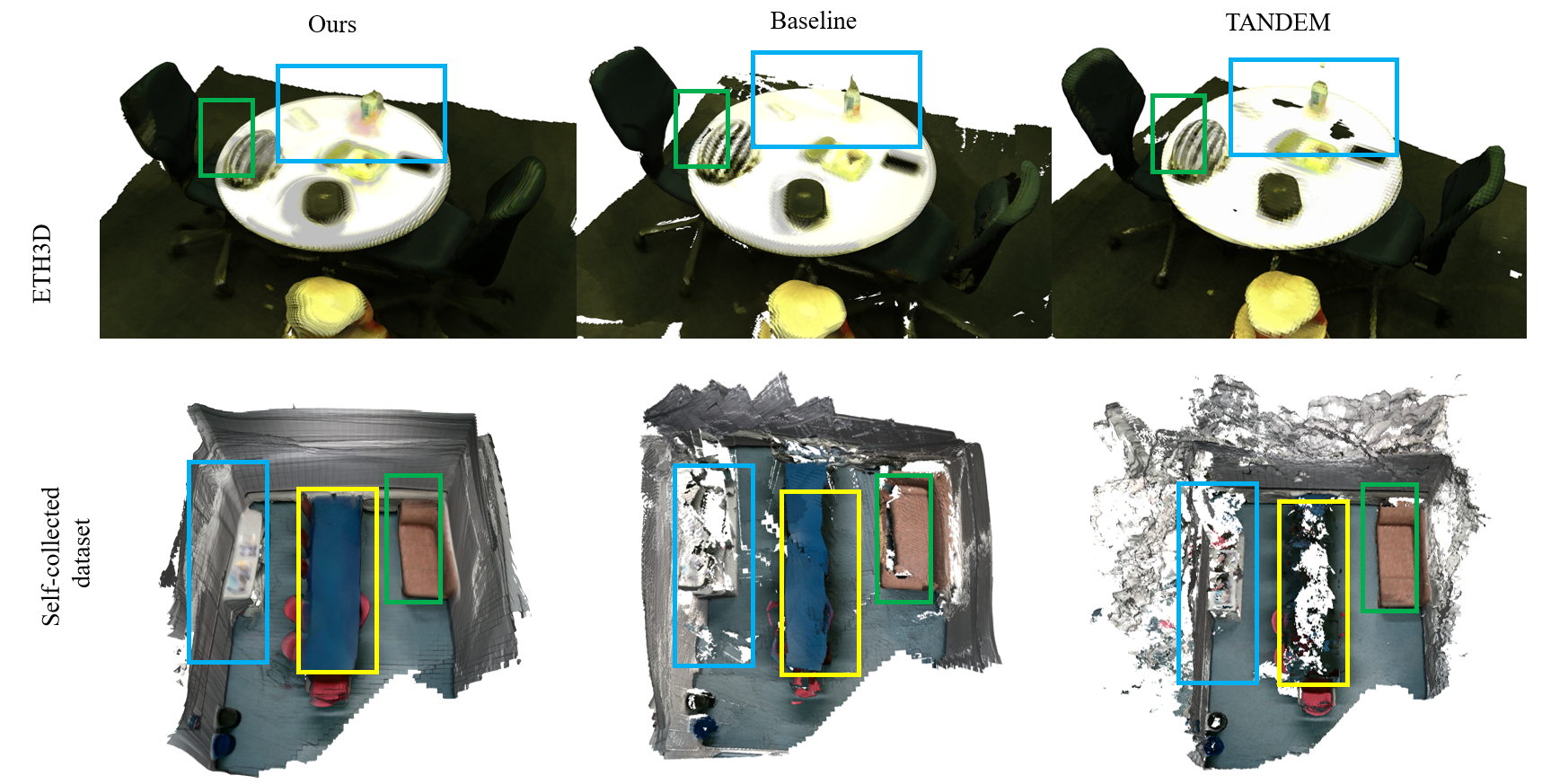}
	\vspace{-1em}
	\caption{3D reconstruction on ETH3D~\cite{schops2019bad} and our self-collected dataset. Both the baseline system and TANDEM~\cite{koestler2021tandem} suffer from inconsistent geometry and noticeable noise.}
	\label{fig:mesh_other}
	\vspace{-1em}
\end{figure}

\PAR{3D Mesh Evaluations.} 
To evaluate the quality of 3D reconstruction, we examine the online reconstructed mesh generated through incremental TSDF-fusion of the keyframe depth maps. Specifically, we assess the performance on the EuRoC dataset~\cite{Burri25012016}, which provides ground truth point cloud data of entire rooms collected through V1 and V2 sequences. However, the collected image sequences in the dataset do not cover all areas of the room, which may have a negative impact on our evaluations. To address this, we prune the entire point clouds and retain only the points covered by camera frustums during image sequence collection. We uniformly sample 800K points from the reconstructed meshes of different methods for evaluation purposes.  To compensate for the limitation of recovering only up-to-scale mesh for monocular methods, we perform Sim3 alignment with the ground truth point cloud. In contrast, our proposed method can directly reconstruct the 3D mesh with the correct scale, enabling us to only perform SE3 alignment during evaluation.


We present the results of our comparison with visual(-inertial) methods in Table \ref{table:mesh_results}. 
Although Sigma-Fusion\cite{rosinol2022probabilistic} exhibits commendable outcomes, it is burdened by limitations in terms of demanding substantial memory and time resources. Both Kimera \cite{rosinol2020kimera} and TANDEM~\cite{koestler2021tandem} exhibit significant performance drops on the challenging V103 and V203 sequences. Our method, on the other hand, showcases substantial performance improvement over them in both easier and more difficult sequences, demonstrating its robustness in challenging scenarios. SimpleMapping surpasses the baseline method significantly in terms of dense mapping accuracy, illustrating the feasibility and advantages of our SPA-MVSNet.
%
%
Figures~\ref{fig:mesh_euroc} and \ref{fig:mesh_other} showcase the qualitative results of our method and the other methods on the 3D reconstruction of the aforementioned datasets. Our method exhibits better detailed reconstruction and less noise, especially on walls, tables, and other details. Notably, we only trained SPA-MVSNet on ScanNet~\cite{dai2017scannet}, and the excellent performance on different unseen datasets underscores the strong generalization capability of SPA-MVSNet and the robustness of our dense mapping system.

\begin{figure}[htb!]
	\centering
	\includegraphics[width=1.0\columnwidth]{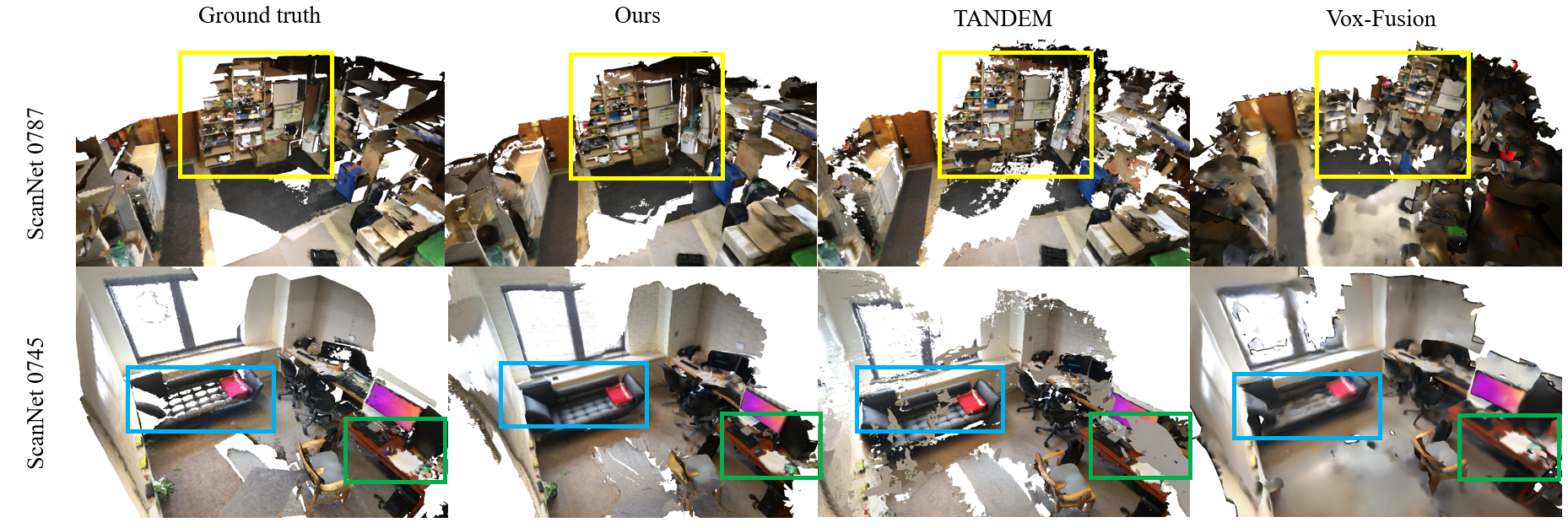}
	\vspace{-2em}
	\caption{3D reconstruction on ScanNet\cite{dai2017scannet} dataset. Vox-Fusion\cite{yang2022vox} tends to produce over-smoothed geometries and experience drift during long-time tracking, resulting in inconsistent reconstruction, as observed in Scene0787.}
	\label{fig:mesh_scannet}
	\vspace{-1em}
\end{figure}

Furthermore, we evaluate our approach against Vox-Fusion~\cite{yang2022vox}, a RGB-D based dense tracking and mapping system using a voxel based neural implicit representation, on ScanNet~\cite{dai2017scannet} test set. Vox-Fusion~\cite{yang2022vox} optimizes feature embeddings in voxels and the weights of a MLP decoder on-the-fly with intensive computation, while ours relying on offline training of the MVS network is much more efficient at inference stage. To ensure fairness, we specifically utilize the released pretrained model on ScanNet \cite{dai2017scannet} for TANDEM~\cite{koestler2021tandem}. As shown in Table~\ref{table:scannet_results} and Figure~\ref{fig:mesh_scannet}, our approach consistently outperforms TANDEM~\cite{koestler2021tandem} and exhibits competitive performance compared to the RGB-D method, Vox-Fusion~\cite{yang2022vox}.

\begin{table}[htb!]
	\centering
	\caption{Mesh evaluation on EuRoC~\cite{Burri25012016} V1 and V2 of mapping systems. 
	$-$ indicates no mesh generated.}
    \label{table:mesh_results}
	\resizebox{1.0\columnwidth}{!}{
	\begin{tabular}{llccccccc}
    	\toprule
        &   & V101   & V102   & V103   & V201   & V202   & V203   & Avg. \\
        \midrule
        \multicolumn{1}{l}{\multirow{5}{*}{Acc [cm]$\downarrow $}} & Kimera \cite{rosinol2020kimera}      & 15.19 & 13.34 & 15.31 & 20.46 & 19.84 & 20.85 & 17.50  \\  
        \multicolumn{1}{l}{}                         & Sigma-Fusion \cite{rosinol2022probabilistic}    & 11.37  & 85.65  & 15.60  & {8.85}  & $-$  & 8.23  & 25.94   \\ 
        \multicolumn{1}{l}{}    & TANDEM\cite{koestler2021tandem}      & \textbf{4.17} & 9.70 & 30.63 & \textbf{6.91} & \textbf{9.17} & 21.03 & 13.60  \\  
        \multicolumn{1}{l}{}                         & Baseline    & 15.41  & 14.09  & 33.33  & 13.95  & 15.58  & 7.98  & 16.72   \\ 
        \multicolumn{1}{l}{}                         & \textbf{Ours}        &  10.50 &  \textbf{7.70} & \textbf{12.62}  & 11.97  & 12.34  & \textbf{7.04}  & \textbf{10.36}   \\
        \midrule
        \multicolumn{1}{l}{\multirow{5}{*}{Comp [cm]$\downarrow $}} & Kimera\cite{rosinol2020kimera}   & 31.16 & 25.46 & 29.30 & 46.06 & 32.84 & 36.79 & 33.60  \\
        \multicolumn{1}{l}{}                         & Sigma-Fusion \cite{rosinol2022probabilistic}    & 14.23  & {11.14}  & 16.11  & {8.92}  & $-$  & 14.94  & 13.06   \\
        \multicolumn{1}{l}{}            & TANDEM\cite{koestler2021tandem}      & 14.68 & 16.18 & 78.41 & 10.86 & 22.56 & 39.40 & 30.35  \\ 
        \multicolumn{1}{l}{}                         & Baseline    & 13.05  & 10.66  & 12.13  & 11.94  & 14.13  & 7.82  & 11.62   \\  
        \multicolumn{1}{l}{}                         & \textbf{Ours}        & \textbf{10.50}  & \textbf{9.73}  & \textbf{7.76}  & \textbf{8.90}  & \textbf{10.40}  & \textbf{6.77}  & \textbf{9.01}   \\ 
        \midrule
        \multicolumn{1}{l}{\multirow{5}{*}{Recall [\%]$\uparrow $}}  & Kimera\cite{rosinol2020kimera}     & 23.99 & 19.64  & 14.42 & 12.74  & 13.33  & 12.72  & 16.14   \\ 
        \multicolumn{1}{l}{}                         & Sigma-Fusion \cite{rosinol2022probabilistic}    & {49.53}  & 49.12  & 25.36 & 51.70  & $-$  & 43.49  & 43.84  \\ 
        \multicolumn{1}{l}{} & TANDEM\cite{koestler2021tandem}      & 40.75 & 32.53  & 9.84  & 42.68  & 27.40  & 17.82  & 28.50   \\ 
        \multicolumn{1}{l}{}                         & Baseline    & \textbf{50.37}  & 44.41  & 35.38 & 52.07  & 44.10  & 50.53  & 46.14   \\  
        \multicolumn{1}{l}{}                         & \textbf{Ours}        & 49.18  & \textbf{51.33}  & \textbf{46.49}  & \textbf{59.43}  & \textbf{52.12}  & \textbf{51.87}  & \textbf{51.74}   \\
        \midrule
        \multicolumn{1}{l}{\multirow{5}{*}{F-score [\%]$\uparrow $}}  & Kimera\cite{rosinol2020kimera}      & 27.64 & 23.56 & 17.84 & 16.30 & 16.64 & 15.01 & 19.50  \\ 
        \multicolumn{1}{l}{}                         & Sigma-Fusion \cite{rosinol2022probabilistic}    & 48.36  & {35.54}  & 27.60  & 50.71  & $-$  & 43.89  & 41.22   \\
        \multicolumn{1}{l}{} & TANDEM\cite{koestler2021tandem}      & \textbf{52.96} & 37.56 & 12.61 & 49.45 & 34.68 & 21.09 & 34.72  \\ 
        \multicolumn{1}{l}{}                         & Baseline    & 49.02  & 41.01  & 31.65  & 48.99  & 42.15  & 48.93  & 43.63   \\  
        \multicolumn{1}{l}{}                         & \textbf{Ours}        & 48.38  & \textbf{47.19}  & \textbf{40.96}  & \textbf{55.09}  & \textbf{48.21}  & \textbf{51.19}  & \textbf{48.50}   \\
        \bottomrule
    \end{tabular}
    }
\end{table}

\begin{table}[htb!]
	\centering
	\caption{Mesh evaluation on ScanNet~\cite{dai2017scannet}. Best results are highlighted as \textbf{first} and \underline{second}. Note that Vox-Fusion~\cite{yang2022vox} takes RGB-D inputs.}
    \label{table:scannet_results}
	\resizebox{1.0\columnwidth}{!}{
	\begin{tabular}{llccccccc}
    	\toprule
        &   & 0736   & 0745   & 0757   & 0786   & 0787   & 0788   & Avg. \\
        \midrule
        \multicolumn{1}{l}{\multirow{3}{*}{Acc [cm]$\downarrow $}} 
        & TANDEM \cite{koestler2021tandem}      & {12.17} & 22.35 & 24.24 & 33.18 & \underline{11.28} & 23.35 & 21.10 \\
        \multicolumn{1}{l}{} & Vox-Fusion \cite{yang2022vox}  & \textbf{2.91}  & \textbf{11.56}  & \textbf{4.82}  & \textbf{9.12}  & 15.43  & \textbf{8.15}  & \textbf{8.66}\\ 
        \multicolumn{1}{l}{}   & \textbf{Ours (Mono)}        &  \underline{9.21} &  \underline{13.36} & \underline{7.05}  & \underline{12.31}  & \textbf{9.86}  & \underline{12.24}  & \underline{10.67}   \\
        \midrule
        \multicolumn{1}{l}{\multirow{3}{*}{Comp [cm]$\downarrow $}} 
        & TANDEM \cite{koestler2021tandem}      & \textbf{12.65} & \underline{13.46} & 53.13 & 181.7 & \underline{9.51} & 28.59 &  49.84\\
        \multicolumn{1}{l}{} & Vox-Fusion \cite{yang2022vox}  & 18.81  & 34.56  & \textbf{6.97}  & \underline{19.23}  & 31.19  & \underline{13.80}  & \underline{20.76}\\  
        \multicolumn{1}{l}{}   & \textbf{Ours (Mono)}   &  \underline{14.31} &  \textbf{7.80} & \underline{22.04}  & \textbf{15.04}  & \textbf{8.50}  & \textbf{9.98}  &  \textbf{12.95}  \\
        \midrule
        \multicolumn{1}{l}{\multirow{3}{*}{Recall [\%]$\uparrow $}}  & TANDEM \cite{koestler2021tandem}      & \underline{42.69} & \underline{47.57} & 18.22 & 10.02 & \underline{48.31} & 16.77 &  30.60 \\
        \multicolumn{1}{l}{} & Vox-Fusion \cite{yang2022vox}  & \textbf{47.22}  & 40.43  & \textbf{59.57}  & \underline{35.06}  & 34.47  & \underline{41.28}  &  \underline{43.00}\\ 
        \multicolumn{1}{l}{}   & \textbf{Ours (Mono)}        &  38.85 &  \textbf{49.07} & \underline{43.26}  & \textbf{41.75}  & \textbf{58.27}  & \textbf{57.59}  &  {\textbf{48.13}}\\
        \midrule
        \multicolumn{1}{l}{\multirow{3}{*}{F-score [\%]$\uparrow $}}   & TANDEM \cite{koestler2021tandem}      & 38.72 & 44.20 & 18.37 & 11.34 & \underline{40.56} & 16.62 &  28.30\\
        \multicolumn{1}{l}{} & Vox-Fusion \cite{yang2022vox}  & \textbf{60.93}  & \textbf{51.36}  & \textbf{61.44}  & \textbf{42.42}  & 39.33  & \underline{45.37}  & \textbf{50.14}\\ 
        \multicolumn{1}{l}{}   & \textbf{Ours (Mono)}        &  \underline{45.76} &  \underline{46.54} & \underline{48.47}  & \underline{40.25}  & \textbf{51.96}  & \textbf{48.05}  &  \underline{46.84}  \\
        \bottomrule
    \end{tabular}
    }
    \vspace{-1em}
\end{table}

\begin{table}[htb!]
    \caption{Ablation study. We evaluate depth estimation on ScanNet~\cite{dai2017scannet}, exploring both the network architecture (top block above the double line) and the robustness against noise in sparse depth (bottom block). DGCV denotes depth prior guided cost volume and CV cost volume. Noise parameters in the bottom block include noise standard deviation in depth $\sigma_d$[m], pixel coordinates $\sigma_{uv}$[pixel], translation $\sigma_{t}$[m], and orientation $\sigma_{R}$[\textdegree] when sampling the sparse points. Default configurations are underlined and italicized.}
    \vspace{-2em}
    \begin{center}
    \setlength{\tabcolsep}{0.3em} 
    \resizebox{1.0\columnwidth}{!}{  
        \begin{tabular}{lccccc}
            \toprule
            & \multicolumn{5}{c}{\textbf{Depth Evaluation}}  \\
            \cmidrule(lr){2-6}
            & Abs Diff [m]$\downarrow$ & Sq Rel [\%]$\downarrow$ & RMSE [m]$\downarrow$ & $\delta < 1.05 [\%]\uparrow$ & $\delta < 1.25 [\%]\uparrow$ \\
            \midrule
            SimpleRecon\cite{sayed2022simplerecon} & 0.0893 & 0.0127 & 0.1478 & 72.84 & 98.05 \\
            Our single-view depth completion & 0.2358 & 0.0777 & 0.3463  & 35.86 & 81.94 \\
            \midrule
            Baseline w/ DGCV, $\lambda = 0.020$ & 0.1401 & 0.0250 & 0.1998 & 50.13 & 94.89 \\
            Baseline w/ DGCV, $\lambda = 0.040$ & 0.0984 & 0.0148 & 0.1602 & 65.33 & 97.39  \\
            Baseline w/ DGCV, $\lambda = 0.075$ & 0.1044 & 0.0160 & 0.1647 & 64.88 & 97.22  \\
            \midrule
            \makecell[l]{Baseline w/ DGCV, 4 feature maps in Enc.} & 0.0753 & 0.0104 & 0.1344 & 77.58 & 98.30  \\
            \makecell[l]{Baseline w/ DGCV, 5 feature maps in Enc.} & 0.0715 & 0.0099 & 0.1295 & 79.43 & 98.46  \\
            Baseline w/ CV, 5 feature maps in Enc. & 0.0774 & 0.0108 & 0.1362 & 76.58 & 98.33 \\
            \makecell[l]{Baseline w/ DGCV, 5 feature maps in Dec.(\textbf{Ours})} & \underline{\emph{0.0696}} & \underline{\emph{0.0088}} &\underline{\emph{0.1215}} & \underline{\emph{79.60}} & \underline{\emph{98.53}}  \\
            Baseline w/ CV, 5 feature maps in Dec. & 0.0793 & 0.0105 
            & 0.1361 & 75.03 & 98.15  \\
            \makecell[l]{Baseline w/ DGCV, 5 feature maps in Enc. and Dec.} & 0.0716 & 0.0098 & 0.1292 & 79.37 & 98.34  \\
            Baseline w/ CV, 5 feature maps in Enc. and Dec. & 0.0820 & 0.0114 & 0.1424 & 74.95 & 98.20 \\
            \midrule
            \midrule
            Ours w/ sparse points 20 & 0.0896 & 0.0124 & 0.1461 & 71.60 & 98.01 \\ 
            Ours w/ sparse points 50 & 0.0804 & 0.0108 & 0.1348 & 74.94 & 98.24 \\ 
            Ours w/ sparse points 100 & 0.0768 & 0.0107 & 0.1341 & 76.81 & 98.35 \\ 
            Ours w/ sparse points 200 & 0.0709 & 0.0090 & 0.1232  & 79.02  & 98.50 \\ 
            Ours w/ sparse points 250 & \underline{\emph{0.0696}} & \underline{\emph{0.0088}} &\underline{\emph{0.1215}} & \underline{\emph{79.60}} & \underline{\emph{98.53}} \\ 
            Ours w/ sparse points 300 & 0.0695 & 0.0093 & 0.1249  & 80.00 & 98.53 \\  
            \midrule
            \makecell[l]{Ours w/ 250 points,  $\sigma_d= 0.45$, $\sigma_{uv} = 3.0$, \\ \qquad \qquad \qquad \qquad $\sigma_{t} = 0.005$, $\sigma_{R} = 0.010$} &  \underline{\emph{0.0696}} & \underline{\emph{0.0088}} &\underline{\emph{0.1215}} & \underline{\emph{79.60}} & \underline{\emph{98.53}} \\ 
            \makecell[l]{Ours w/ 150 points, $\sigma_d = 0.6$, $\sigma_{uv} = 5.0$, \\ \qquad \qquad \qquad \qquad  $\sigma_{t} = 0.02$, $\sigma_{R} = 0.015$}  & 0.0763 & 0.0106 & 0.1329 & 76.83  & 98.32 \\ 
            \makecell[l]{Ours w/ 200 points, $\sigma_d = 0.8$, $\sigma_{uv} = 6.0$,  \\ \qquad \qquad \qquad \qquad $\sigma_{t} = 0.03$, $\sigma_{R} = 0.020$}  & 0.0825 & 0.0118 & 0.1386 & 72.97  & 98.02 \\ 
            \makecell[l]{Ours w/ 100 points, $\sigma_d = 0.8$, $\sigma_{uv} = 6.0$, \\ \qquad \qquad \qquad \qquad $\sigma_{t} = 0.03$, $\sigma_{R} = 0.020$}  & 0.0850 & 0.0123 & 0.1422 & 72.31 & 98.01  \\
            \makecell[l]{Ours w/ 50 points,  $\sigma_d =0.8$, $\sigma_{uv} =6.0$, \\ \qquad \qquad \qquad \qquad $\sigma_{t} =0.05$, $\sigma_{R} =0.040$} & 0.0897  & 0.0126 & 0.1440 & 69.98 & 97.86 \\
            \bottomrule
        \end{tabular}
    }
    \end{center}
    \vspace{-2.5em}
    \label{table:ablations}
\end{table}

\subsection{Ablations}
\label{sec:ablations}
We conduct an ablation study of our proposed SPA-MVSNet on the test split of ScanNet~\cite{dai2017scannet}. Our study examines the network architecture design of SPA-MVSNet as well as the impact of noise levels on sparse points during the inference stage. The results of the ablation study for depth estimation metrics are presented in Table~\ref{table:ablations}. 


\PAR{Baseline.} We investigate the depth predictions of the MVS backbone, SimpleRecon~\cite{sayed2022simplerecon}, and our single-view depth completion network. The experiments reveal that the dense depth predictions from depth completion are notably inferior to those from SimpleRecon~\cite{sayed2022simplerecon}. Nevertheless, the depth predictions from the single-view depth completion network can serve as a useful prior for our SPA-MVSNet, as elaborated in the following ablation analyses.

\PAR{Depth Prior Guided Cost Volume Formulation.} We explore the integration of single-view depth prior for the purpose of cost volume formulation in our SPA-MVSNet. Specifically, we generate 64 uniformly spaced depth hypotheses around the depth prior with equal intervals in the cost volume, and analyze the impact of varying the interval. Table~\ref{table:ablations} shows the evaluation results of depth prior guided cost volume (DGCV) with intervals of $0.02$m, $0.04$m, and $0.075$m. Our experiments demonstrate that a fixed interval of $0.04$m achieves the best cost volume formulation. 
However, simply introducing depth priors into the cost volume formulation does not necessarily benefit depth predictions. This is due to the fact that the MVS regularized network lacks information about the distance between the surfaces assumed in the formulation.

\PAR{Deep Depth Feature Integration.} 
We further enhance the MVS 2D CNN regularization network by integrating hierarchical depth features from the decoder of the single-view depth completion network into MVS encoder-decoder architecture (see Fig.~\ref{fig:system_overview}). 
%
Maintaining a cost volume interval of $0.04$m, we explore the impact of integrated depth feature map quantity and different concatenation strategies on the MVS encoder-decoder pipeline. Our findings demonstrate that utilizing a single-view depth prior in combination with cost volume formulation and 2D CNN regularization can greatly enhance the accuracy of depth prediction.
%
%
Additionally, we validate the efficacy of the depth prior guided cost volume by contrasting it with the same network architecture employing the conventional cost volume, which uniformly covers the 3D space from minimum to maximum depth using a fixed interval. The outcomes clearly indicate that the depth prior guided cost volume yields significant improvement.

\PAR{The number of Sparse Points.} During training, we simulate a sparse depth map by sampling a fixed number of sparse points (\ie, 250), analogous to the landmarks generated by a VIO system.
However, in practice, the number of valid sparse points projected on the sparse depth map can vary. In our experiments, we observe that the number of valid VIO sparse points varies from 100 to 300 in typical scenes. Therefore, evaluating the performance of SPA-MVSNet with varying numbers of sparse points is vital.
We follow the same strategy for sampling noisy sparse points as during training (see Section~\ref{sec:network training}). As shown in Table \ref{table:ablations}, the proposed SPA-MVSNet method demonstrates a high degree of generalization across a range of sparse point densities, from 20 to 300. 
%
The results show that having fewer sparse points with relatively accurate depth values does not significantly affect the depth prediction performance.

\PAR{Robustness to Noise in Sparse Points.} To evaluate the noise tolerance of our SPA-MVSNet during inference, we conduct experiments by testing the network using parameter settings that affect the noise levels in the generated sparse depth maps. During training, we maintain a constant number of sparse points at 250, with noise generation parameters set at $\sigma_d=0.45$m, $\sigma_{uv}=3.0$ pixels, $\sigma_{t}=0.005$m, and $\sigma_{R}=0.01$\textdegree. As illustrated in Table \ref{table:ablations}, even under reasonably higher noise levels, SPA-MVSNet exhibits minor performance degradation and does not suffer from disastrous performance reduction.


\subsection{Runtime Performance}
SimpleMapping can process images at an approximate rate of 20 frames per second when executed on a desktop equipped with an NVIDIA GeForce RTX 3070 graphics card with 8GB of VRAM and an Intel i5-10600 CPU. The complete dense mapping system includes tracking, bundle adjustment (in LocalMapping), MVS depth prediction, and TSDF-Fusion. During the evaluation of running time, visualization is disabled. Table~\ref{table:runtime} presents the average runtimes for different modules of SimpleMapping and TANDEM~\cite{koestler2021tandem} on EuRoC~\cite{Burri25012016} V1 and V2, both executed on the same hardware device. The tracking and bundle adjustment modules are processed on the CPU, while depth prediction and fusion modules are processed in parallel on the GPU. Notably, TANDEM~\cite{koestler2021tandem} provides an additional runtime evaluation scheme that preloads sensor data and deploys the tracking thread on the GPU, thereby improving the real-time efficiency to an average of 14.5 FPS, which is still slower than ours.

While examining scene reconstruction methods based on neural implicit representations, we observe these methods demand significant memory resources and involve time-intensive operations, leading to poor real-time performance. For example, when applying Vox-Fusion~\cite{yang2022vox} to the ScanNet dataset~\cite{dai2017scannet}, each frame's average processing time for tracking amounts to 2.32 seconds. Similarly, the optimization process for the neural implicit representation requires 2.25 seconds. These evaluations are conducted using an NVIDIA A40 graphics card, indicating a significant gap from the desired real-time capability.

\begin{table}[htb!]
    \caption{Runtime evaluation of mapping systems on EuRoC~\cite{Burri25012016} V1 and V2. 
    We present the averaged per-keyframe runtime for each module and per-frame runtime for the whole process. SimpleMapping is able to ensure real-time performance, only requiring 55 ms to process one frame.}
    \vspace{-1em}
    \begin{center}
    \resizebox{0.80\columnwidth}{!}{  
        \begin{tabular}{lcc}
            \toprule
            &  TANDEM~\cite{koestler2021tandem} & \textbf{Ours} \\   
            \midrule
            Tracking [ms]& 49.7 & 15.9  \\
            Bundle Adjustment [ms]& 29.0 & 78.0  \\ 
            MVS Depth Prediction [ms] & 223.9 & 138.3 \\
            TSDF-Fusion [ms]& 68.6  & 67.1  \\
            \midrule
            \midrule
            Avg. [ms]& 135.3  (7.4 FPS) & 54.6 (18.3 FPS) \\
            \bottomrule
        \end{tabular}
    }
    \end{center}
    \vspace{-2em}
    \label{table:runtime}
\end{table}

\subsection{Discussion about Limitations}
SimpleMapping faces challenges in low-texture scenarios due to the reliance on sparse visual points. In extreme cases, the VIO can only recover very few sparse points with significant noise, which would lead to implications in the final depth prediction of the SPA-MVSNet. Additionally, SPA-MVSNet may struggle with occluded or highly reflective surfaces, as well as large depth discontinuities.

\section{Conclusion}

We present SimpleMapping, a real-time visual-inertial dense mapping system that delivers pleasing 3D reconstruction with metric scale in an incremental fashion. 
One of our key contributions is the sparse depth aided MVS network (SPA-MVSNet), that utilizes the 6-DoF poses and noisy sparse depth estimated from VIO to recover dense depth maps for VIO reference keyframes. 
%
%
We validate entire dense mapping system on two public datasets and our own collected dataset, showcasing its state-of-the-art dense mapping performance with great robustness, high efficiency, and remarkable generalization capabilities.
%
SimpleMapping provides both real-time 6-DoF pose estimation and dense scene reconstruction with realistic and detailed geometry, which makes it well-suited for augmented reality (AR) and virtual reality (VR) applications.
%
%
For future work, we plan to expand the current VIO tracking into a full SLAM system with loop closure correction. This will enhance the consistency of dense mapping in large-scale scenarios.

\acknowledgments{
}
This work is partially supported by Munich Center for Machine Learning (MCML), Germany.

\bibliographystyle{abbrv-doi}

\bibliography{egbib}
\end{document}